%% file: main.tex
\documentclass[10pt,twocolumn,letterpaper]{article}

\usepackage[pagenumbers]{cvpr} 

\input{preamble}
\definecolor{cvprblue}{rgb}{0.21,0.49,0.74}
\usepackage[pagebackref,breaklinks,colorlinks,allcolors=cvprblue]{hyperref}

\def\confName{CVPR}
\def\confYear{2027}

\newcommand{\alias}{EgoStudy\xspace}

\title{EgoStudy: How Does Scaling Egocentric Data Affect VLA and WAM Pretraining?}

\author{
Puzhen Yuan$^{1,2,*}$\quad
Yanjiang Guo$^{1,2,*}$\quad
Yichen Liu$^{1,2,*}$\quad
Xiang Zhu$^{1,2}$\quad
Qingzhou Lu$^{1,2}$\\[3pt]
Chenxin Liu$^{1,2}$\quad
Tun Ban$^{2,3}$\quad
Haodong Yan$^{4}$\quad
Zhide Zhong$^{4}$\quad
Jianyu Chen$^{1,2,\dagger}$\\[6pt]
{\normalsize $^{1}$Tsinghua University\quad $^{2}$Shanghai Qi Zhi Institute\quad $^{3}$Shanghai Jiao Tong University\quad $^{4}$HKUST (GZ)}\\[3pt]
{\small $^{*}$Equal contribution\quad $^{\dagger}$Corresponding author}\\[4pt]
{\small Project page: \url{https://puzhenyuan.github.io/EgoStudy-website}}
}

\begin{document}
\makeatletter
\g@addto@macro\@maketitle{\input{sec/teaser}}
\makeatother
\maketitle
\input{sec/0_abstract}    
\input{sec/1_introduction}
\input{sec/2_related_works}
\input{sec/3_study_design}
\input{sec/4_experiments}
\input{sec/5_conclusion}
\input{sec/6_statement}
{
    \small
    \bibliographystyle{ieeenat_fullname_unsrt}
    \bibliography{main}
}


\end{document}

%% file: preamble.tex
\usepackage{float} 

%% file: sec/teaser.tex
\vspace{-10pt}
\begingroup
\centering
\begin{minipage}{\textwidth}
    \centering
    \captionsetup{hypcap=false}
    \includegraphics[width=0.95\linewidth]{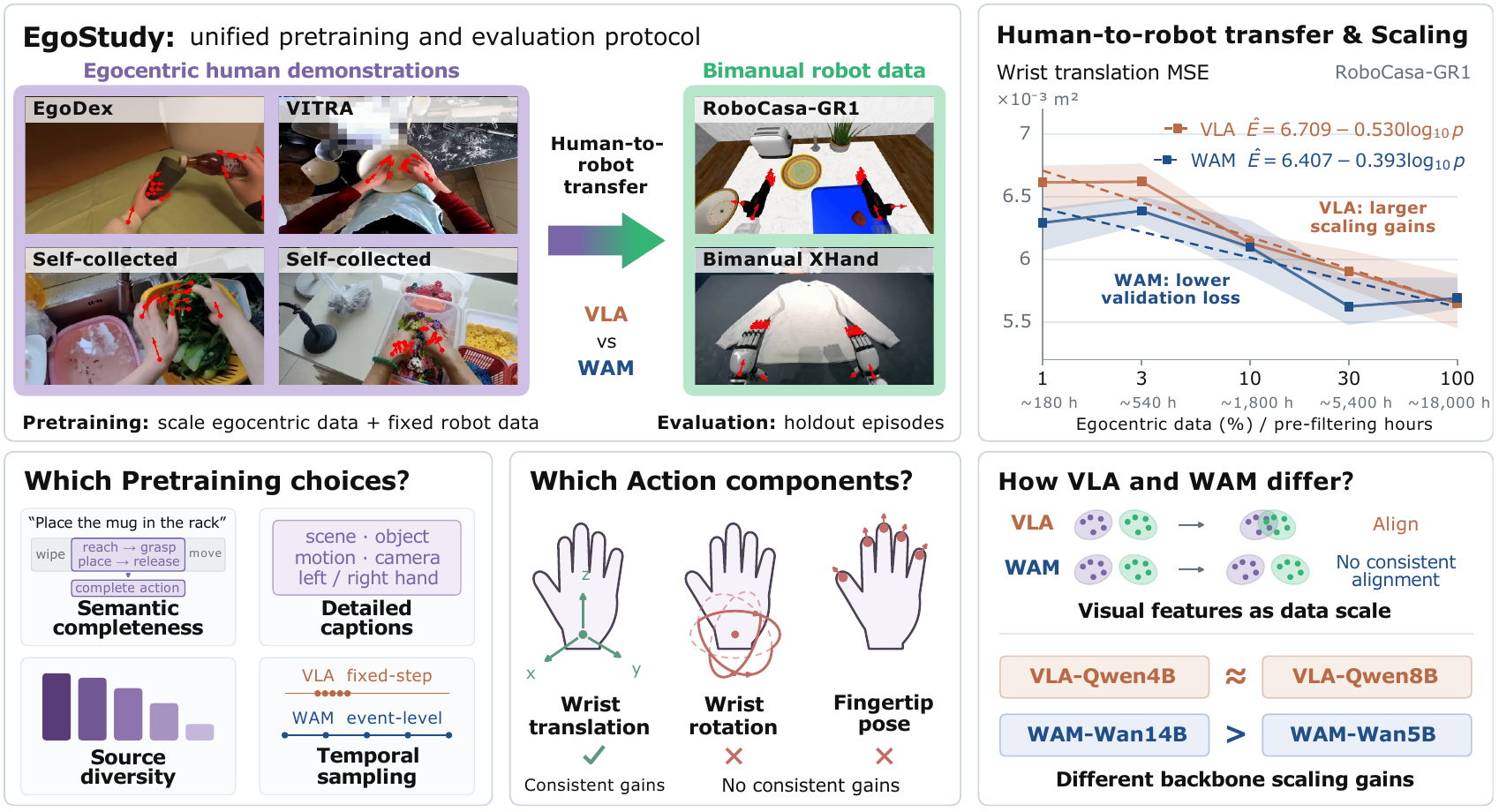}
    \captionof{figure}{\textbf{Overview of \alias.}
    We conduct a controlled study by scaling egocentric human data while keeping robot training exposure fixed (top left).
    Increasing human data reduces wrist translation error for both VLAs and WAMs (top right).
    The bottom panels show how scaling behavior varies with pretraining choices, across action components, and between VLAs and WAMs.}
    \label{fig:teaser}
\end{minipage}
\par\vspace{16pt}
\endgroup

%% file: sec/0_abstract.tex
\begin{abstract}

Egocentric human manipulation data offer a scalable resource for training dexterous robot policies. However, how effectively robots can learn and benefit from these data remains unclear due to the embodiment gap between humans and robots.
We present EgoStudy, a systematic, controlled study of egocentric data for robot learning. Holding the robot dataset fixed, we vary the scale and diversity of egocentric data, data processing strategies, and co-training configurations. We investigate three questions: (1) Which data processing and pretraining choices most effectively exploit egocentric data? (2) Which components of robot action prediction benefit most from scaling these data? (3) How do scaling effects differ between vision-language-action models (VLAs) and world-action models (WAMs)?
Our experiments yield three main findings. 
First, data diversity, alignment between trajectories and detailed language annotations are key factors in effective transfer. 
Second, scaling egocentric data yields larger gains in predicting end-effector translation than in predicting fine-grained finger movements. 
Third, WAMs perform better when egocentric data are limited, whereas VLAs exhibit greater gains as the amount of egocentric data increases.
Together, these findings provide practical guidance on processing and using egocentric data to train VLAs and WAMs, while clarifying the benefits and limitations of scaling these data.

\end{abstract}

%% file: sec/1_introduction.tex
\section{Introduction}
\label{sec:intro}

Learning general-purpose dexterous manipulation requires experience spanning diverse objects, environments, and interaction patterns.
Robot demonstrations provide direct supervision for control, but collecting them at scale requires substantial hardware resources and effort for operation and maintenance.
Egocentric human videos offer an attractive source of manipulation experience: people perform a broad range of everyday activities, exposing models to interactions that are expensive to reproduce through robot data collection.
Recent efforts have made this experience increasingly accessible through hand-motion annotations~\cite{yu2025dyn,potamias2025wilor,hoque2026egodex}, creating opportunities to pretrain Vision--Language--Action Models (VLAs)~\cite{kim2024openvla,physicalintelligence2025pi05,pmlr-v229-zitkovich23a} and World--Action Models (WAMs)~\cite{ye2026dreamzero,guo2026videopredictionpolicy2} with growing amounts of human experience.
However, limited attention~\cite{li2025vitra,guo2026videopredictionpolicy2,zhu2026harp} has been given to the fundamental questions about scaling egocentric data for robot foundation model pretraining:
\begin{itemize}
    \item Which data processing and pretraining choices most effectively exploit egocentric data?
    \item Which components of robot action prediction benefit most from scaling these data?
    \item How do egocentric scaling effects differ between VLAs and WAMs? 
\end{itemize}


Answering these questions requires examining more than data volume alone.
Existing studies have shown that egocentric human videos can improve robot policy learning, but their conclusions are difficult to compare. Different studies typically use different robot datasets, action representations, model architectures, optimization budgets, and evaluation protocols. Consequently, it remains unclear whether an observed improvement comes from additional human experience, a particular data-processing choice, or a model-specific training recipe. 
Moreover, most evaluations report an aggregate action loss, leaving open which components of dexterous control actually benefit from egocentric pretraining and whether these offline gains translate to closed-loop control.

Studying VLAs and WAMs together is particularly informative because they learn from different pretrained priors and objectives.
Representative VLAs adapt vision-language representations for action prediction~\cite{kim2024openvla,black2024pi_0,kareer2025emergence,zhu2026harp,liu2026gala,chen2026villax}, while video-based WAMs combine action learning with future visual prediction~\cite{yuan2026fastwam,hu2024video,guo2026videopredictionpolicy2,guo2024pad}.
These differences suggest that the same egocentric experience may benefit the two model families in different ways.

To investigate these questions, we conduct a systematic empirical study named \alias, a unified pretraining and evaluation protocol for VLAs and WAMs.
\alias combines egocentric human demonstrations and bimanual robot data through a shared dexterous action interface describing wrist motion and fingertip poses, providing common action targets across embodiments and supporting comparisons between model families.
We vary only the scale and construction of the egocentric human data, including episode curation, source diversity, caption detail, and temporal sampling, while holding robot training data and exposure fixed.
We assess human-to-robot transfer using a common evaluation protocol with held-out samples, component-wise action metrics, and validation loss on robot data.
Finally, standardized robot posttraining and evaluation on target robot domain ensure that pretraining improvements translate into better deployment outcomes.



The results reveal that the benefits of egocentric pretraining are selective rather than uniform.
Controlled comparisons in \alias show that VLAs show larger gains from scaling egocentric data, and WAMs generally attain lower robot-domain validation loss.
Semantically complete episodes, detailed language captions, diverse egocentric data, fixed-step sampling for VLA and event-level sampling for WAM pretraining all contribute to lower robot validation loss and stronger human-to-robot transfer.
Scaling egocentric data improves wrist-motion prediction more than finger-motion prediction. 
Furthermore, we compare how the visual representations of VLAs and WAMs evolve as egocentric data increases and how their gains vary with backbone capacity, providing a basis for explaining why the two model families scale differently.
Together, these findings provide practical guidance and empirical insights for training VLAs and WAMs with human videos, while revealing the capability gains and limitations of scaling egocentric data.

Our contributions are as follows:
\begin{itemize}
    \item \textbf{Pretraining protocol}. We introduce \alias, which includes a unified pretraining and evaluation protocol to support systematic study of egocentric human-to-robot transfer across VLAs and WAMs.
    \item \textbf{Comprehensive study}. We conduct comprehensive experiments of data scale, semantic completeness, caption detail, source diversity, temporal sampling, action components and backbone capability, identifying how these choices affect robot foundation model pretraining.
    \item \textbf{Practical guidance}. We derive practical recommendations for improving human-to-robot transfer, outline expected limitations, and offer insights into the performance discrepancies between VLA and WAM scaling.
\end{itemize}






%% file: sec/2_related_works.tex
\section{Related Work}

\subsection{Robot Foundation Models}
\label{sec:foundation_models}

Vision--language--action models adapt pretrained vision--language backbones to predict robot actions from observations and language instructions~\cite{zhang2026vlm4vla,guo2025irevla,guo2026vlaw,huang2026breakinglockin}.
RT-2~\cite{pmlr-v229-zitkovich23a} and OpenVLA~\cite{kim2024openvla} predict discrete action tokens, while GR00T N1~\cite{gr00tn1_2025} and $\pi_{0.5}$~\cite{physicalintelligence2025pi05} use an action expert to generate continuous action sequences.
Recent work also explores how to connect perception and control, either by separating high-level visual reasoning from low-level action generation~\cite{cui2025openhelix,zhang2025hirt} or by training a shared model for multimodal understanding and future prediction~\cite{zhang2025upvla,hu2026bagelvla}.

Video-based policies also learn from how a scene changes during manipulation. The VPP family~\cite{hu2024video,guo2026videopredictionpolicy2} and GE-Act~\cite{agibot2025genie} use predictive video features to generate actions. DreamZero~\cite{ye2026dreamzero}, Fast-WAM~\cite{yuan2026fastwam}, OpenWAM~\cite{wang2026openwam}, and LingBot-VA~\cite{li2026lingbotva} couple video and action prediction during training, with different choices about how the two streams interact and whether to generate future frames at inference. We use StarVLA~\cite{community2026starvla} and Fast-WAM~\cite{yuan2026fastwam} to compare these two model families in the same dexterous action space.

\subsection{Learning from Egocentric Videos}
\label{sec:learn_from_ego}

Egocentric demonstrations provide two forms of supervision for robot manipulation. First, tracked human hand motion supplies action targets for dexterous imitation~\cite{wang2024dexcap,yang2025egovla} and for co-training on human and robot demonstrations~\cite{kareer2024egomimic,li2026aceego}. VITRA~\cite{li2025vitra} and EgoScale~\cite{zheng2026egoscale} pair egocentric observations with hand trajectories for policy pretraining and report improved downstream manipulation as the amount of human data increases. Kareer et al.~\cite{kareer2025emergence} find that diverse robot pretraining improves transfer from human video, while HumanScale~\cite{ma2026humanscale} shows that curated human demonstrations can outperform an equal amount of robot pretraining data. Even with accurate hand tracks, differences in how humans and robots move can limit transfer~\cite{li2026everyday,wu2026atomego,zhang2026egoforesight}.

Second, video records the effects of an action on the scene, such as an object moving as a hand grasps it. OpenWAM~\cite{wang2026openwam}, Open-HumanWAM~\cite{ma2026openhumanwam}, and GE-Act 2.0~\cite{agibot2026geact} use human manipulation footage to train visual prediction alongside or before action learning. In a comparison of prediction targets, EgoWAM~\cite{li2026egowam} reports stronger transfer on bimanual tasks when predicting visual features or 3D motion instead of pixels. Ego4WAM~\cite{sun2026ego4wam} studies how human--robot alignment, task diversity, and action supervision affect transfer within a fixed WAM. We compare VLAs and WAMs at fixed robot exposure and measure how scaling human data affects wrist translation, wrist rotation, and fingertip pose.

%% file: sec/3_study_design.tex
\section{Study Design}
\label{sec:study_design}

\alias provides a shared pretraining pipeline and matched evaluation protocols to enable controlled comparisons of egocentric human-to-robot transfer in VLAs and WAMs.
The study examines which data choices improve robot action prediction, how the gains differ between wrist and finger motion, and the scaling difference between VLAs and WAMs.

\begin{figure}[!t]
    \centering
    \includegraphics[width=\linewidth]{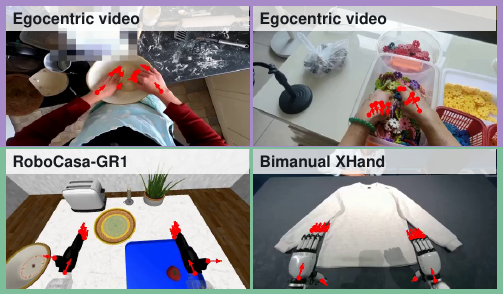}
    \caption{Red points and arrows indicate positions and orientations (omitting 2 axes for clarity), respectively, in shared dexterous action space and different embodiments.}
    \label{fig:shared_action_examples}
\end{figure}

\subsection{Model Architecture}
\label{sec:study_models}

The VLA pairs a pretrained Qwen3-VL-4B-Instruct vision--language backbone with a GR00T-style action expert~\cite{community2026starvla,gr00tn1_2025}.
The backbone encodes the current camera views and instruction. Given these features and the current geometric state, the action expert predicts continuous actions.

The WAM pairs a pretrained video backbone Wan2.1-I2V-14B-480P with an action diffusion transformer for joint video--action prediction~\cite{yuan2026fastwam}.
Action tokens attend to video latents as the model jointly predicts future video and actions from the initial observation, instruction, and state.

During pretraining, both models predict 32-step chunks in the same 108-dimensional action space.
Each hand contributes a wrist pose and five fingertip poses, each encoded as a 3D position and a 6D rotation~\cite{zhou2019rotation}.
Let $(p_w(t),R_w(t))$ and $(p_f(t),R_f(t))$ denote the world poses of a wrist and one of its fingertips.
We denote the anchor time by $0$ and the rotation from initial camera axes to world axes by $R_{c,0}$.
The wrist targets at time $t$ are
\begin{align}
    \Delta p_w(t) &= R_{c,0}^{\top}\bigl(p_w(t)-p_w(0)\bigr),\\
    \Delta R_w(t) &= R_{c,0}^{\top} R_w(t) R_w(0)^{\top} R_{c,0}.
\end{align}
Wrist translation $\Delta p_w(t)$ captures displacement from the initial wrist position, expressed in the initial camera frame.
Wrist rotation $\Delta R_w(t)$ captures the change in wrist orientation in the same frame.
The fingertip targets at that time are
\begin{align}
    p_{f\mid w}(t) &= R_w(t)^{\top}\bigl(p_f(t)-p_w(t)\bigr),\\
    R_{f\mid w}(t) &= R_w(t)^{\top}R_f(t).
\end{align}
Fingertip poses describe the position and orientation of each fingertip relative to its wrist at the target step, capturing hand articulation separately from wrist motion.
Each rotation matrix is encoded by its first two columns to obtain the 6D representation.
\Cref{sec:component_scaling} examines how transfer from egocentric data differs across these action components.
\Cref{fig:shared_action_examples} illustrates the evaluation data and shared dexterous action representation.

\begin{figure}[!t]
    \centering
    \includegraphics[width=\linewidth]{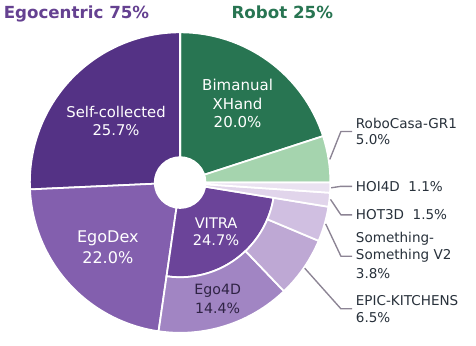}
    \caption{\textbf{Pretraining samples allocation.}
    Egocentric datasets receive 75\% of samples, with probabilities proportional to the square root of their filtered durations; VITRA aggregates three subsets.
    The sampling probabilities of Bimanual XHand (20\%) and RoboCasa-GR1 (5\%) are fixed across experiments.}
    \label{fig:data_sampling}
    \par\vspace{6pt}
    \begin{minipage}{\linewidth}
        \centering
        \captionof{table}{Post-filtering duration of the pretraining data (hours). The egocentric pool is curated from about 18,000 hours of raw video.}
        \label{tab:data_duration}
        \small
        \setlength{\tabcolsep}{3pt}
        \begin{tabular*}{\linewidth}{@{\extracolsep{\fill}}llr@{}}
            \toprule
            \textbf{Category} & \textbf{Dataset} & \textbf{Hours} \\
            \midrule
            Egocentric & EgoDex & 633.831 \\
            & VITRA (subtotal) & 347.217 \\
            & \quad Ego4D & 272.855 \\
            & \quad EPIC-KITCHENS & 55.808 \\
            & \quad Something-Something V2 & 18.554 \\
            & HOT3D & 2.936 \\
            & HOI4D & 1.525 \\
            & Self-collected & 867.414 \\
            \midrule
            Robot & Bimanual XHand & 272.676 \\
            & RoboCasa-GR1 & 80.837 \\
            \midrule
            \multicolumn{2}{@{}l}{\textbf{Total}} & \textbf{2,206.436} \\
            \bottomrule
        \end{tabular*}
    \end{minipage}
\end{figure}

Both models use AdamW with cosine learning-rate decay and matched batch sizes and update counts.
Masked action flow matching supervises both models, with an additional video flow-matching loss for the WAM.
Supplementary details the model training settings and action representation.

\subsection{Data Composition and Processing}
\label{sec:study_data}

The human training pool combines self-collected demonstrations with hand-annotated clips from the public egocentric datasets EgoDex~\cite{hoque2026egodex}, HOI4D~\cite{liu2022hoi4d}, HOT3D~\cite{banerjee2024hot3d}, and VITRA~\cite{li2025vitra}.
VITRA contributes clips from Ego4D~\cite{grauman2022ego4d}, EPIC-KITCHENS~\cite{damen2021epickitchens}, and Something-Something V2~\cite{goyal2017somethingsomething}.
Robot demonstrations come from diverse tabletop tasks in self-collected Bimanual XHand and simulated kitchens in RoboCasa-GR1~\cite{nasiriany2024robocasa}.

A common episode format stores synchronized RGB frames, camera poses, and hand poses at 20\,Hz, together with a language caption.
Human hand poses come from annotations or tracking, and robot hand poses from forward kinematics. The curation pipeline aligns both to a common coordinate convention.
Source-specific segmentation produces semantically complete action clips, which a vision--language model annotates with detailed captions.
Filtering further removes clips with unusable video, hand tracks, or camera trajectories. Validity masks exclude invalid pose coordinates, and action normalization uses fixed per-source statistics.
Supplementary details the segmentation and captioning pipelines for human and robot demonstrations.

The main experiment applies the same data fraction to all egocentric sources while fixing each robot source's training pool and sampling probability.
Within the human mixture, weights are proportional to the square root of each source's eligible training time count, tempering the influence of larger datasets.
At a fixed global batch size and update count, each robot source contributes the same expected number of samples, allowing us to compare robot learning across human data scales.

\begin{figure*}[t]
    \centering
    \begin{subfigure}[t]{0.495\textwidth}
        \includegraphics[width=\linewidth]{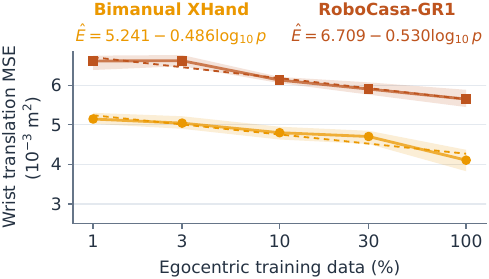}
        \caption{VLA robot validation.}
        \label{fig:transfer_vla_loss}
    \end{subfigure}\hfill
    \begin{subfigure}[t]{0.495\textwidth}
        \includegraphics[width=\linewidth]{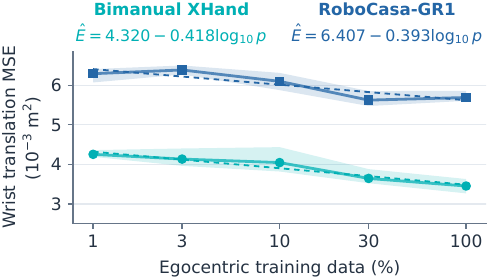}
        \caption{WAM robot validation.}
        \label{fig:transfer_wam_loss}
    \end{subfigure}

    \caption{\textbf{Does scaling egocentric data improve robot learning?} 
    We pretrain the VLA (left) and WAM (right) with 1\%, 3\%, 10\%, 30\%, and 100\% of the egocentric human data while keeping robot training exposure and the optimization budget fixed. We evaluate wrist translation MSE on held-out trajectories from Bimanual XHand and RoboCasa-GR1. All four model–domain pairs show a consistent downward trend, indicating effective human-to-robot transfer in \alias.}
    \label{fig:transfer_scaling}
\end{figure*}

\subsection{Experimental Protocol and Metrics}
\label{sec:study_metrics}

Offline evaluation measures masked mean squared error on denoised actions in unnormalized coordinates.
We report separate errors for wrist translation, wrist rotation, and fingertip poses.
Following the action representation in \cref{sec:study_models}, we evaluate denoised predictions after reversing action normalization.
Define
\begin{equation}
    \operatorname{MSE}_m(\hat x,x)
    = \frac{\sum_j m_j(\hat x_j-x_j)^2}{\sum_j m_j},
\end{equation}
where $j$ indexes scalar coordinates across all samples and target steps in each robot holdout, and $m_j$ indicates validity.
The component losses are
\begin{equation}
    \begin{aligned}
        \mathcal L_{\mathrm{trans}}
        &= \operatorname{MSE}_m(\widehat{\Delta p}_w,\Delta p_w),\\
        \mathcal L_{\mathrm{rot}}
        &= \operatorname{MSE}_m(\hat r_w,r_w),\\
        \mathcal L_{\mathrm{finger}}
        &= \operatorname{MSE}_m(\hat q_f,q_f),
    \end{aligned}
\end{equation}
where $r_w=\rho_6(\Delta R_w)$, $q_f=[p_{f\mid w};\,\rho_6(R_{f\mid w})]$, and $\rho_6$ concatenates the first two columns of a rotation matrix.
The fingertip loss jointly averages position and rotation coordinates. We report the mean value on 3 training seeds and error bars.

For closed-loop evaluation, we adapt each pretrained policy to RoboCasa-GR1 under a matched posttraining and control protocol.
We posttrain on the official dataset only with fixed-step sampling from the pretrained checkpoint, and use a pretrained neural IK to convert fingertip poses to hand joints for deployment.
For Robocasa, we report task success rate over 10 trials per task, resulting 240 trials overall.
Supplementary material gives the posttraining hyperparameters, action interface, controller, and rollout settings.

%% file: sec/4_experiments.tex
\section{Experiments and Analysis}
\label{sec:experiments}

\begin{figure*}[t]
    \centering
    \begin{subfigure}[t]{0.485\textwidth}
        \includegraphics[width=\linewidth]{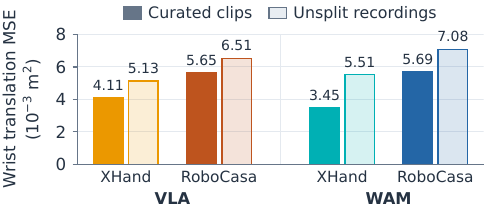}
        \caption{Episode semantic completeness.}
        \label{fig:ablation_complete}
    \end{subfigure}\hfill
    \begin{subfigure}[t]{0.485\textwidth}
        \includegraphics[width=\linewidth]{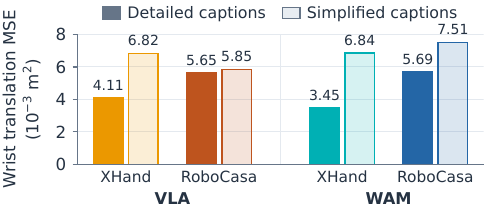}
        \caption{Caption detail.}
        \label{fig:ablation_caption}
    \end{subfigure}
    \medskip
    \begin{subfigure}[t]{0.485\textwidth}
        \includegraphics[width=\linewidth]{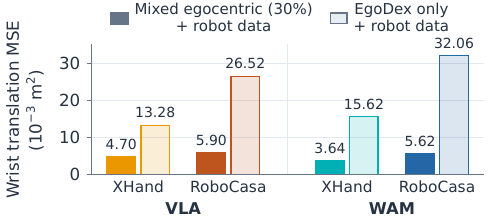}
        \caption{Source diversity.}
        \label{fig:ablation_diversity}
    \end{subfigure}\hfill
    \begin{subfigure}[t]{0.485\textwidth}
        \includegraphics[width=\linewidth]{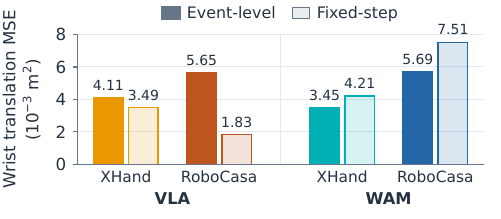}
        \caption{Sampling method.}
        \label{fig:ablation_sampler}
    \end{subfigure}
    \caption{\textbf{Which data processing and co-training designs improve transfer?}
    Complete episodes, detailed captions, and the dataset diversity yield lower wrist translation error on both robot holdouts for both model families. This indicates that data diversity and alignment between trajectories and detailed language annotations are key factors in effective transfer.
    Moreover, fixed-step sampling yields lower error for the VLA, and event-level sampling yields lower error for the WAM.}
    \label{fig:pretrain_ablations}
\end{figure*}

Using the protocol in \cref{sec:study_design}, we first assess human-to-robot transfer in \alias through held-out robot action prediction and RoboCasa-GR1 task success rates (\cref{sec:transfer_results}).
The remaining sections address the three proposed questions:
\begin{itemize}
    \item \Cref{sec:transfer_ablations} tests which data processing and pretraining choices improve human-to-robot transfer.
    \item \Cref{sec:component_scaling} examine which component of action capabilities improve as human data scale.
    \item \Cref{sec:model_scaling} investigates why VLAs and WAMs exhibit different scaling behavior by comparing their visual representations and examining the role of backbone capacity.
\end{itemize}

\newcommand{\EgoResultPanel}[2][25mm]{%
    \begingroup
    \setlength{\fboxsep}{5pt}%
    \fbox{\begin{minipage}[c][#1][c]{\dimexpr\linewidth-2\fboxsep-2\fboxrule\relax}
        \centering\footnotesize
        \textit{Data pending}\par\smallskip
        #2\par
    \end{minipage}}%
    \endgroup
}

\subsection{Human-to-Robot Transfer}
\label{sec:transfer_results}

The scaling sweep tests whether additional human experience improves robot learning under a fixed robot data budget.
Each scale changes the eligible human training pool while preserving robot pools, their sampling probabilities, and the optimization budget, as described in \cref{sec:study_data}.
\Cref{fig:transfer_scaling} compares the default VLA and WAM configurations.

Each semilog fit $\hat{E}(p)=a+b\log_{10}p$ summarizes the change in mean wrist translation MSE per tenfold increase in the human data percentage $p$.
From $1\%$ to $100\%$, the reported error falls by $20.2\%$ for the VLA and $18.8\%$ for WAM on Bimanual XHand, and by $14.6\%$ and $9.6\%$, respectively, on RoboCasa-GR1.
Mean Bimanual XHand error decreases at each successive fraction for both models.
Overall, VLAs exhibit more pronounced scaling gains, and WAMs achieve lower robot validation loss and higher downstream performance.

\begin{figure}[H]
    \centering
    \includegraphics[width=\linewidth]{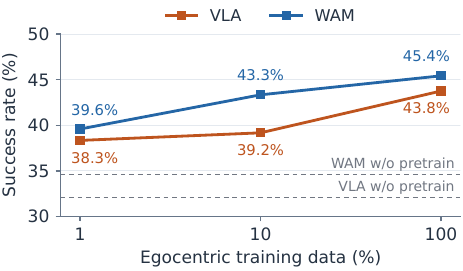}
    \caption{\textbf{Do pretraining gains carry over to robot control?}
    RoboCasa-GR1 task success increases after posttraining for both model families, consistent with the decrease in pretraining validation loss.
    Gray dashed lines mark the corresponding baselines without pretraining.}
    \label{fig:transfer_success}
\end{figure}

Under the matched RoboCasa-GR1 evaluation protocol in \cref{sec:study_metrics}, the VLA success rate rises from $38.3\%$ to $43.8\%$ and the WAM rate from $39.6\%$ to $45.4\%$ as the egocentric fraction increases from $1\%$ to $100\%$ (\cref{fig:transfer_success}).
At the three fractions with rollout results, lower RoboCasa-GR1 validation error coincides with higher posttraining success for both models, providing a basis for using robot-domain validation loss to characterize pretraining performance.

\subsection{Processing and Pretraining Choices}
\label{sec:transfer_ablations}

Four ablations test which egocentric pretraining choices improve transfer to the robot domains, with each ablation changing one pretraining factor at a time within a model family.
These factors include episode completeness, caption detail, source diversity, and temporal sampling.
The comparisons match optimization budgets and robot exposure within each model family.
\Cref{fig:pretrain_ablations} reports wrist translation MSE on the shared robot holdout.

\noindent\textbf{Episode semantic completeness.}
In this experiment, we compare the impact of splitting data into semantically complete action segments on egocentric pretraining.
Unsplit recording ablation preserves the original video duration while simply adding detailed annotations.
At matched training steps on the full human pool, curated clips yield lower validation error than unsplit recordings for both model families in both robot domains.
Segmentation and filtering retain complete actions with a recognizable goal and observable completion.
This gives the model a clearer correspondence between the instruction and the sampled motion during joint pretraining on human and robot data (See supplementary material for visualization).

\noindent\textbf{Detailed Caption.}
The simplified-caption ablation retains the trajectories and pose targets and replaces task instructions with a simplified caption during training and holdout evaluation.
The simplified caption only contains two words: one verb for an action, one word for an object.
Both models benefit from detailed captions on the diverse real-robot XHand holdout and Robocasa-GR1 simulation holdout.

\noindent\textbf{Source diversity.}
After filtering, EgoDex contains slightly more than $30\%$ of the multi-source egocentric pool.
In this ablation experiment, we use only EgoDex and robot data for pretraining, comparing against that which utilizes 30\% mixed-source egocentric data. 
For both the VLA and WAM, EgoDex-only training yields higher wrist translation MSE on both robot holdouts, reaching $2.8$--$5.7$ times the corresponding mixed-source reference values in \cref{fig:ablation_diversity}.
At comparable human data volumes, the multi-source mixture provides broader coverage of actions, objects, and environments for robot foundation model pretraining.
The mixed condition also retains robot demonstrations, so the observed gap reflects changes in both egocentric source diversity and robot supervision.

\begin{figure*}[t]
    \centering
    \includegraphics[width=\textwidth]{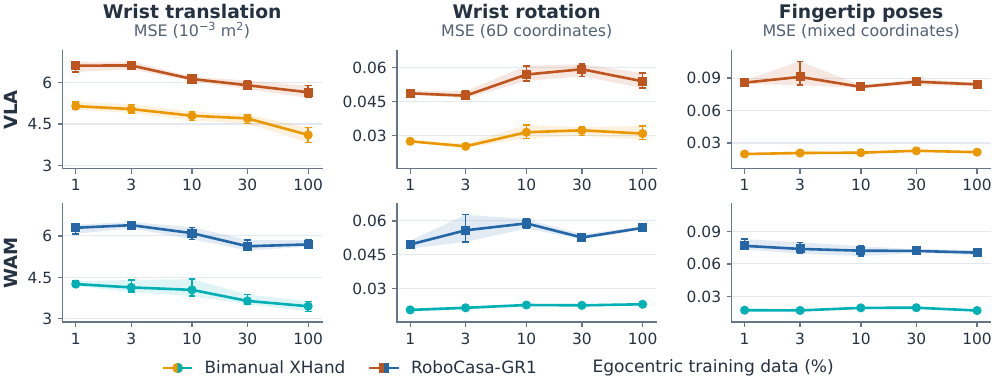}
    \caption{\textbf{Which action components benefit most from scaling egocentric data?} We measure mean squared error (MSE) separately for wrist translation, wrist rotation, and fingertip pose on high-quality, held-out robot trajectories. Across both robot domains and model families, \textbf{wrist translation MSE generally decreases} as egocentric data scale, while \textbf{wrist rotation and fingertip pose errors show no consistent decline}. We hypothesize that due to human–robot embodiment gap and errors in estimating actions limit transfer primarily to large-scale motions such as wrist translation, with less benefit for the finger movements and wrist rotations.}
    \label{fig:component_scaling}
\end{figure*}

\noindent\textbf{Sampling method.}
VPP2~\cite{guo2026videopredictionpolicy2} and WALL-WM~\cite{li2026wallwm} use semantically coherent events during pretraining to improve generalization across manipulation tasks.
Motivated by these studies, we test whether sampling across a complete event during pretraining benefits both model families.
Event-level sampling distributes action targets uniformly across the episode, whereas fixed-step sampling selects the same number of targets at a constant stride. We use the same episodes, captions, and evaluation timestamps in both conditions.
On both robot holdouts, event-level sampling yields lower wrist translation MSE for the WAM, while fixed-step sampling yields lower MSE for the VLA (\cref{fig:ablation_sampler}).
One possible explanation is that the WAM benefits from observing how an event unfolds, while a fixed stride gives the VLA a consistent time interval for predicting displacement.
For the scaling experiments, we use event-level sampling for both models to keep the temporal sampling protocol matched (\cref{fig:transfer_scaling}).

\subsection{Transfer Across Action Components}
\label{sec:component_scaling}

The second question in \cref{sec:intro} asks which robot prediction capabilities improve as egocentric data scale.
\Cref{fig:component_scaling} separates wrist translation, wrist rotation, and fingertip pose error at $1\%$, $3\%$, $10\%$, $30\%$, and $100\%$ of the human pool under fixed robot exposure.
Across both robot domains, the reported wrist translation errors show an overall decrease for both model families, with larger relative reductions for the VLA.
Wrist rotation and fingertip pose errors show no common improvement across models and domains.
We hypothesize that errors in action labels estimated from egocentric video obscure small changes in wrist orientation and fingertip pose more readily than larger wrist displacements.
During contact, even a small pose error can change which part of a finger touches an object, and the adjustment needed to maintain contact can differ between human and robot hands.
These factors may limit how much additional egocentric data helps predict the fine motions required for dexterous manipulation.


\begin{figure*}[t]
    \centering
    \includegraphics[width=\linewidth]{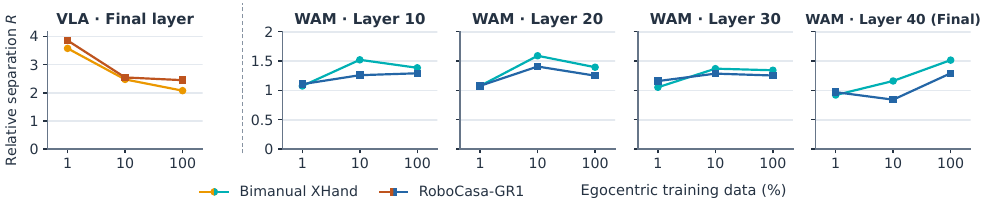}
    \caption{\textbf{How do VLA and WAM visual features change with scale?}
    As egocentric data scale, relative separation $R$ between human and robot features decreases in both robot domains for the VLA but shows no consistent decrease across the WAM layers examined.
    We hypothesize that human and robot features align more closely in our VLA because it only needs to output actions in a shared action space. Our WAM must also predict future video, which may encourage it to retain visual differences between human and robot hands.}
    \label{fig:vla_features}
\end{figure*}

\begin{figure*}[t]
    \centering
    \includegraphics[width=\linewidth]{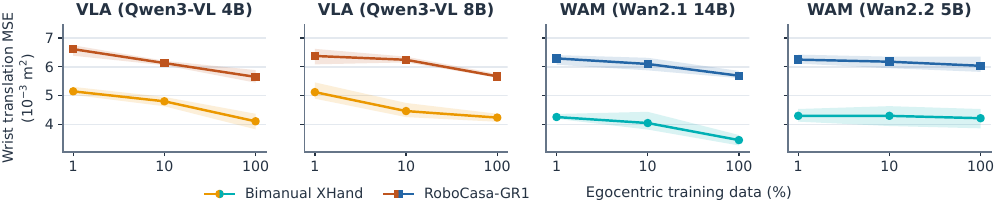}
    \caption{\textbf{How does backbone choice affect scaling gains between VLA and WAM?}
    The wrist translation error decreases to the same extent across different VLAs, indicating that the 4B backbone is already capable of effectively extracting features shared between the egocentric and robot data. When the backbone of WAM is changed to 5B, its human-to-robot capability declines indicating that WAM relies more heavily on a powerful video backbone to assimilate egocentric data.}
    \label{fig:backbone_scaling}
\end{figure*}

\subsection{How VLAs and WAMs Scale Differently}
\label{sec:model_scaling}

The default VLA in the preceding experiments uses StarVLA with a Qwen3-VL-4B-Instruct backbone and a GR00T-style action expert.
The default WAM uses the joint Fast-WAM architecture with a Wan2.1-I2V-14B-480P video backbone.
In \Cref{fig:transfer_scaling}, using action validation loss as the metric, VLA demonstrates a more pronounced human-to-robot transfer effect on both robot domains, whereas WAM achieves a lower validation loss.

We examine how visual representations change with egocentric data scale and vary the backbones to investigate what causes the differences in human-to-robot transfer behavior between VLA and WAM.

\noindent\textbf{Cross-embodiment feature alignment.}
For each final checkpoint, we mean-pool visual tokens from the same observation frames and $\ell_2$-normalize each resulting feature.
We probe the final VLA layer with native camera views and WAM layers 10, 20, 30, and 40 (final) with the main camera frame.
Relative separation $R$ divides the distance between the egocentric and robot feature means by their pooled within-domain dispersion.
Across the three data fractions shown, $R$ decreases for the VLA on both robot domains (\cref{fig:vla_features}).
In contrast, the feature distributions across the various layers of WAM do not converge as the egodata scale increases—unlike what is observed with VLA.

\noindent\textbf{Backbone capability.}
We change the VLA backbone to Qwen3-VL-8B-Instruct and WAM backbone to Wan2.2-TI2V-5B to test the influence of backbone capability.

Wrist translation error falls with scale for both the Qwen3-VL 4B and Qwen3-VL 8B VLAs (\cref{fig:backbone_scaling}).
The reduction in relative feature separation is consistent with an action objective that favors shared motion cues and discounts embodiment-specific appearance.
Robot demonstrations provide task-specific action supervision, and egocentric demonstrations expand the coverage of shared motions. The observed scaling gains suggest that a 4B backbone can already exploit this structure.

Wan2.1 wrist translation error falls with scale in both domains just like VLAs. However, a less capable Wan2.2 backbone struggles to translate scaled egodata into improved robot action prediction performance (\cref{fig:backbone_scaling}).
The configurations differ only in the video backbone, suggesting that WAMs rely more on powerful pretrained backbones than VLAs.

These findings reveal a previously underestimated distinction between VLA and WAM: 
VLA models embodiment-invariant actions, which can be learned effectively using smaller models by focusing on the commonalities within action data—thereby bypassing the need to model embodiment- or scene-specific variations and making the data easier to scale. 
In contrast, WAM benefits more from a superior video base model, depending on the video backbone's capacity to assimilate egocentric video data, thus enhancing video prediction capabilities yields more significant scaling gains—with actions merely being extracted from the video. 
However, pixel-level prediction tends to distract WAM by forcing it to focus on the specific embodiment, which may ultimately limit its scaling potential compared to VLA.
Moreover, in terms of the trend toward egocentric scaling, video priors may serve a function similar to that of egocentric pretraining for robot foundation models, as both provide prior knowledge regarding diverse environments, motions, and objects.

%% file: sec/5_conclusion.tex
\section{Conclusion}
\label{sec:conclusion}

\alias brings egocentric human and bimanual robot demonstrations into a shared dexterous action space for controlled VLA and WAM pretraining and evaluation, enabling effective human-to-robot transfer.
We conduct comprehensive experiments of data scale, semantic completeness, caption detail, source diversity, temporal sampling, action components and backbone capability, identifying how these choices affect robot foundation model pretraining.
First, data diversity, alignment between trajectories and detailed language annotations are key factors in effective transfer. 
Second, scaling egocentric data yields larger gains in predicting end-effector translation than in predicting fine-grained finger movements. 
Third, WAMs perform better when egocentric data are limited, whereas VLAs exhibit greater gains as the amount of egocentric data increases.
Experiments involving visual features and backbone variations indicate that the scaling of VLA may stem from the modeling of shared actions, whereas the scaling of WAM may depend on the video backbone's capacity to assimilate egocentric video data—with actions merely being extracted from the video prediction.
Together, these findings reveal the capability gains and limitations of scaling egocentric data.

%% file: sec/6_statement.tex
\section{LLM Usage Statement}
\label{sec:statement}

We used large language models (LLMs) to assist with manuscript polishing, figure refinement, and experimental implementation. LLMs were not used to generate or develop the core research ideas, hypotheses, or methodological contributions. All scientific content, experimental results, and conclusions were independently verified by the authors, who take full responsibility for the work.